\documentclass[10pt,twocolumn,letterpaper]{article} 

\usepackage{avss}
\usepackage{times}
\usepackage{epsfig}
\usepackage{graphicx}
\usepackage{amsmath}
\usepackage{amssymb}
\usepackage{booktabs}
\avssfinalcopy % *** Uncomment this line for the final submission

\def\avssPaperID{63} % *** Enter the AVSS Paper ID here

\ifavssfinal\fi
\begin{document}

%%%%%%%%% TITLE
\title{Deep-Temporal LSTM for Daily Living Action Recognition}

\author{Srijan Das, Michal Koperski, Francois Bremond\\
INRIA, Sophia Antipolis\\
2004 Rte des Lucioles, 06902, Valbonne, France\\
{\tt\small name.surname@inria.fr}
% For a paper whose authors are all at the same institution, 
% omit the following lines up until the closing ``}''.
% Additional authors and addresses can be added with ``\and'', 
% just like the second author.
% To save space, use either the email address or home page, not both
\and
Gianpiero Francesca\\
Toyota Motor Europe\\
Hoge Wei 33, B - 1930 Zaventem\\
{\tt\small gianpiero.francesca@toyota-europe.com}
}

\maketitle
% \thispagestyle{empty}

%%%%%%%%% ABSTRACT
\begin{abstract}
   In this paper, we propose to improve the traditional use of RNNs by employing a many to many model for video classification. We analyze the importance of modeling spatial layout and temporal encoding for daily living action recognition. Many RGB methods focus only on short term temporal information obtained from optical flow. Skeleton based methods on the other hand show that modeling long term skeleton  evolution improves action recognition accuracy. In this work, we propose a deep-temporal LSTM architecture which extends standard LSTM and allows better encoding of temporal information. In addition, we propose to fuse 3D skeleton geometry with deep static appearance. We validate our approach on public available CAD60, MSRDailyActivity3D and NTU-RGB+D, achieving competitive performance as compared to the state-of-the art.
\end{abstract}

%%%%%%%%% BODY TEXT
\section{Introduction}
In this work we focus on solving problem of daily living action recognition. This problem facilitates 
many applications such as: video surveillance, patient monitoring and robotics. The problem is challenging due to complicated nature of human actions such as: pose, motion, appearance variation or occlusions.

Holistic RGB approaches focus on computing hand-crafted or deep features (eg. CNN). Such methods usually model short term temporal information using optical flow. Long term temporal information is either ignored or modeled using sequence classifiers such as HMM, CRF or most recently LSTM. 

Introduction of low-cost depth sensors and advancements in skeleton detection algorithms lead to increased research focus on skeleton based action recognition. 3D skeleton information allows to build action recognition methods based on high level features, which are robust to view-point and appearance changes~\cite{gemetricfeaturesWACV2017}. RNNs are a variant of neural nets capable of handling sequential data, applied on this problem to model the dynamics of human motion. Existing work on RNN based action recognition model the long term contextual information in the temporal domain to represent the action dynamics and so on. The traditional many-to-one LSTMs used for video classification takes decision based on the feature obtained at the last time stamp, failing to incorporate the prediction of video class over time in the loss backpropagated. This disables the LSTM to model long-term motion for action classification. \\
%Similarly to holistic methods -- different ways of modeling temporal information were proposed: HMM, CRF, trees, LSTM.
In this paper we propose a many-to-many model for video classification namely, deep-temporal LSTM in which the features extracted and the loss backpropagated is computed over time. We show that this latent representation of video from LSTM leads to better temporal encoding. We also propose to fuse RGB information and skeleton information. We claim that it is especially important in daily-living action recognition, where many actions have similar motion and pose footprint (eg. drinking and taking pills), thus it is very important to model appearance of objects involved in the action accurately.

This paper shows (1) that LSTM can model temporal evolution of activities when the loss is computed over every temporal frames rather than relying on the last time step, (2) deep appearance and motion based features selected from appropriate image region for action classification. In our work, we propose to use late fusion of skeleton based LSTM classifier with appearance based CNN classifier. Both classifiers work independently and we fuse their classification scores to obtain final classification label. In this way, we take advantage of LSTM classifier which is able to capture long term temporal evolution of pose and CNN classifier which focuses on static appearance features. 

We validate our work on 3 public daily-activity datasets: CAD60, MSRDailyActivity3D and NTU-RGB+D. Our experiments show that we obtain competitive results on all the datasets as compared to the state-of-the-art.

%------------------------------------------------------------------------

\section{Related Work}
\par Previous work on human action recognition was centric with the use of dense trajectories~\cite{3} combined with Fisher Vector (FV) aggregation. The introduction of low cost kinect sensors have made it possible to detect the skeleton poses of human body easily which can be exploited to recognize actions as in ~\cite{8a} or 3D trajectories~\cite{koperski:hal-01054949}. 
\par The emergence of Deep Learning in computer vision has improved the results in terms of accuracy of action recognition as they show some promising results~\cite{10}. One of the key point to use deep learning for action recognition is that, it not only focuses on extracting the deep features from CNNs but for instance considers the temporal evolution of these features using the Recurrent Neural Networks (RNNs).\\  
In ~\cite{twostream}, authors have used two stream networks for action recognition. One for appearance features from the RGB frames and one for flow based features from optical flow. They propose to fuse these features in the last convolutional layer rather than fusing them in the softmax layer which further improves their accuracy of the framework. Cheron \etal~\cite{22} have used different parts of the skeleton so as to extract the CNN features from each of them. These features are aggregated with min-max pooling to classify the actions. The authors claim to use the temporal information by taking the difference of these CNN features followed by the max-min aggregation. But this aggregation ignores the temporal modeling of the spatial features. The recent advancement in this field led to the use of 3D CNNs in C3D and I3D~\cite{i3d} for action classification reporting high accuracy. But these networks containing huge number of parameters are difficult to train on small datasets. They also do not address the long term dependencies of the actions.\\
The availability of informative three dimensional human skeleton data led to the use of RNNs which are capable of modeling the dynamics of human motion. Shahroudy \etal~\cite{NTU_RGB+D} proposed the use of stacked LSTMs namely, Deep-LSTM and also, p-LSTM where separate memory cells are dedicated for each body part of an individual. Another variant of LSTM is proposed in~\cite{st-lstm}, where the authors introduce a new gating mechanism within LSTM to learn the reliability of the sequential data. Some studies also reports the use of different types of feature on RNNs as in~\cite{lrcn,gemetricfeaturesWACV2017}. Authors in ~\cite{lrcn} modeled the spatio-temporal relationship by feeding the LSTM with CNN features from fc-6 of VGG network. But this strategy is valid for datasets with very dynamic actions and not applicable on similar based motion characterized actions. Authors in~\cite{gemetricfeaturesWACV2017} have represented 3D skeletons using distance based features and feed them into 3 layer LSTM. They have also proposed joint line distance to be the most discriminative features for action classification. \\ 
From the above discussion, it is clear that action recognition tasks focus on improving appearance and motion based features and temporal features through RNN modeling separately. But both spatial layout and temporal encoding is important to model daily living activities. This is because of the presence of low motion actions like \textit{typing keyboard, relaxing on couch} and so on where spatial layout is important, and similar actions like \textit{drinking water, brushing hair} and so on where temporal encoding is important. Thus we propose to combine features from convolutional network and recurrent network to encode appearance-motion and temporal information together in a model. \\
In this paper, we use body translated joint coordinates from the depth information to find the discriminative dynamics of the actions using 3-layer LSTM followed by a SVM classification for the temporal stream. We show that our LSTM based features can model the actions temporally better that the existing LSTM architectures with similar input sequences. For deep spatial features, we extended ~\cite{22} to produce CNN features considering both the flow and appearance features from different image regions. We employ a feature selection mechanism to use the most informative image-region over the training data for classifying actions. Since temporal information modeling along with encoding spatial layout is an important dimension in action recognition, so we focus on using the fusion of deep spatial features along with temporal information to recognize actions using a late fusion to learn semantic concepts from unimodal features.

\section{Proposed Method}

\subsection{Deep-Temporal LSTM}
\begin{figure}
\begin{center}
   \includegraphics[width=0.8\linewidth]{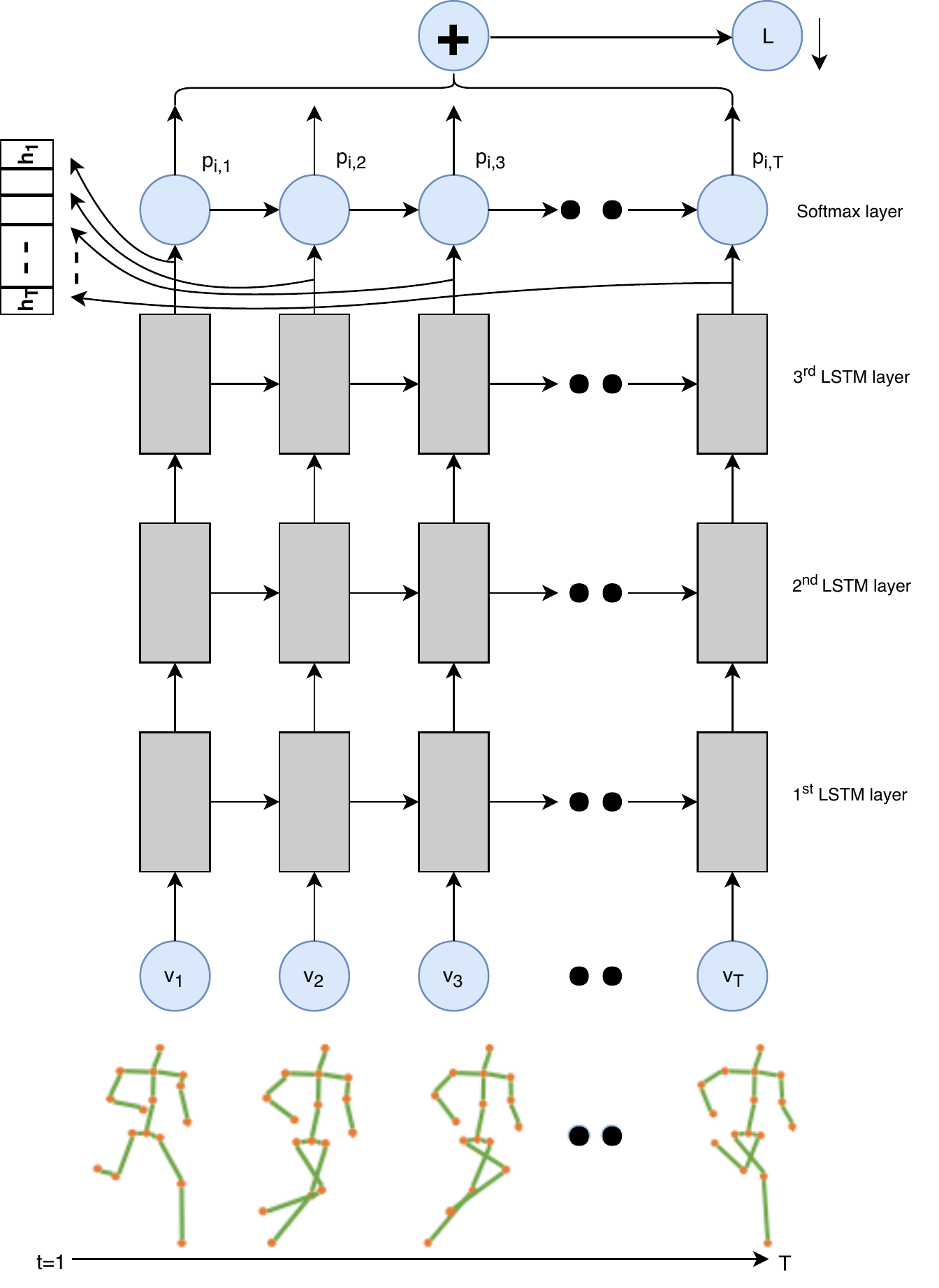} % include this on page two
\end{center}
   \caption{Three-layer stacked LSTM with $t=T$ time steps. The skeleton joint coordinates $v_t$ are input at each time step. $L$ is the loss computed over time and $h$ is the latent vector from last layer LSTM which is input to the linear SVM classifier.}
\label{LSTM_architecture}
\end{figure}

LSTMs being a special kind of RNNs can model the time information as in~\cite{gemetricfeaturesWACV2017}. LSTM mitigates the vanishing gradient problem faced by RNNs by utilizing the gating mechanism over an internal memory cell. The gates enable the LSTM to determine what new information is going to be stored in the next cell state and what old information should be discarded. Such recurrent model receives inputs sequentially and models the information from the seen sequence with a componential hidden state $h_t$: 
\begin{equation}
h_t = f_h(h_{t-1}, v_t;\theta_h)
\end{equation}
where LSTM is our recurrent function $f_h$ with parameters $\theta_h$. We omit the gates from the equations so as to keep the notation simple. The input to the recurrent model is the context vector $v_t$ which is described below. \\
The main focus of the existing methods includes using the RNNs to discover the dynamics and patterns for 3D human action recognition. The sequential nature of the 3D skeleton joints over the times makes the RNN learn the discriminative dynamics of the body.
In this work, we use transformed body pose information on a 3-layer stacked LSTM so as to model the temporal information as shown in fig. ~\ref{LSTM_architecture}. The main reason for stacking LSTM is to allow for greater model complexity, to perform hierarchal processing on large temporal tasks and naturally capture the structure of sequences. A pre-processing step is performed to normalize the 3D skeleton in camera coordinate system as in~\cite{NTU_RGB+D}. The 3D skeleton joint is translated to the $hip-center$ followed by a rotation of the X axis parallel to the 3D vector from "right hip" to the "left hip", and Y axis towards the 3D vector from "spine base" to "spine". At the end, we scale all the 3D joints based on the distance between "spine base" and "spine" joints. Thus the transformed 3D skeleton $v_t$ at time frame $t$ which is represented as $[x_{r,t}, y_{r,t}, z_{r,t}]$ for $r \in $ joints $(J)$ and $(x,y,z)$ being the spatial location of $r^{th}$ joint is input to the LSTM at time stamp $t$. We normalize the time steps in videos by padding with zeros. This is done to keep fixed time steps in LSTM to process a video sample. \\

Traditionally, authors in~\cite{lrcn,gemetricfeaturesWACV2017,NTU_RGB+D} solve action recognition problem as a many to one sequence classification problem. They compute the loss at the last time step of the video  which is backpropagated through time. In this work, we compare the LSTM cell output with the true label of the video at each timestep. In this way we get time-step sources to correct errors in the model (via backpropagation) rather than just for each video (giving rise to the term \textit{deep-temporal}). Thus the cost function of the LSTM for videos is computed by averaging the loss at each frame as follows
\begin{equation}
L = -\frac{1}{N}{\sum\limits_{i=1}^{N}\sum\limits_{t=1}^{T}y_{i}\log(p_{it})}
\end{equation}
where $L$ is categorical cross-entropy computed for $N$ video samples in a batch over $T$ frames, $y$ is the sample label and $p_{it} \in (0,1): \Sigma_t = 1 \forall i,t$ is the prediction for a video. This loss $L$ is back-propagated through time. Here, the LSTM treats each temporal sequence independently as a sample, whose prediction is again determined by the current  and previous gate states. This method provides better performance compared to the minimization of the loss at the last time step only due to better feedback backpropagated through time. So, we extract the latent vectors from every time step of the last layered LSTM using it as a feature extractor. \\
We train deep-temporal LSTM with parameters $\theta_h$ on the input sequence $ V = \{v_t\}$ with loss $L$ resulting in a hidden state representation $h = \{h_t\}$. Each element in $h_t$ is again a latent vector represented as $h_t = \{h_{j,t}\}$, where $j$ is the index over the hidden state dimension. This latent vector constituting $h'_i = h_{r}$ represent the action dynamics at time instant $r \in \{ 1 \cdots T \}$ for $i^{th}$ sample, qualifying it as a representative vector over time. This latent vector $h$ represents a better and more complex representation of the long-term dependencies among the input 3D sequential data. This temporal latent vector $h$ is input to a linear SVM classifier for action classification. The 3-D matrix $H_n = \{h'_1, h'_2, \cdots h'_n\}$ for $n$ training sample is input to the SVM to learn the mapping $\mathbb{X} \rightarrow \mathbb{Y}$, where $h'_n \in \mathbb{X}$ and $y \in \mathbb{Y}$ is a class label. The features $\mathbb{X}$, extracted from trained LSTM are used to learn a classifier: $y = f_{SVM}(h'_n,\alpha)$, where $\alpha$, the parameters of the function $f_{SVM}$.
\vspace{-0.1in}
\subsection{Pose based CNNs}
In ~\cite{22}, the authors have used the concept of two streams for recognizing actions on the different parts of the subject extracted from their skeleton joint information. This inspires us to use the deep features from different body regions of the subject so as to represent their appearance and motion features. The main objective behind using these features is to model the static appearances along with encoding the object information carried while performing the actions. We extend~\cite{22} by invoking deeper networks for feature extraction and employing a feature selection technique to select the best image region involved in the action dataset. \\
\begin{figure}
\begin{center}
   \includegraphics[width=1\linewidth]{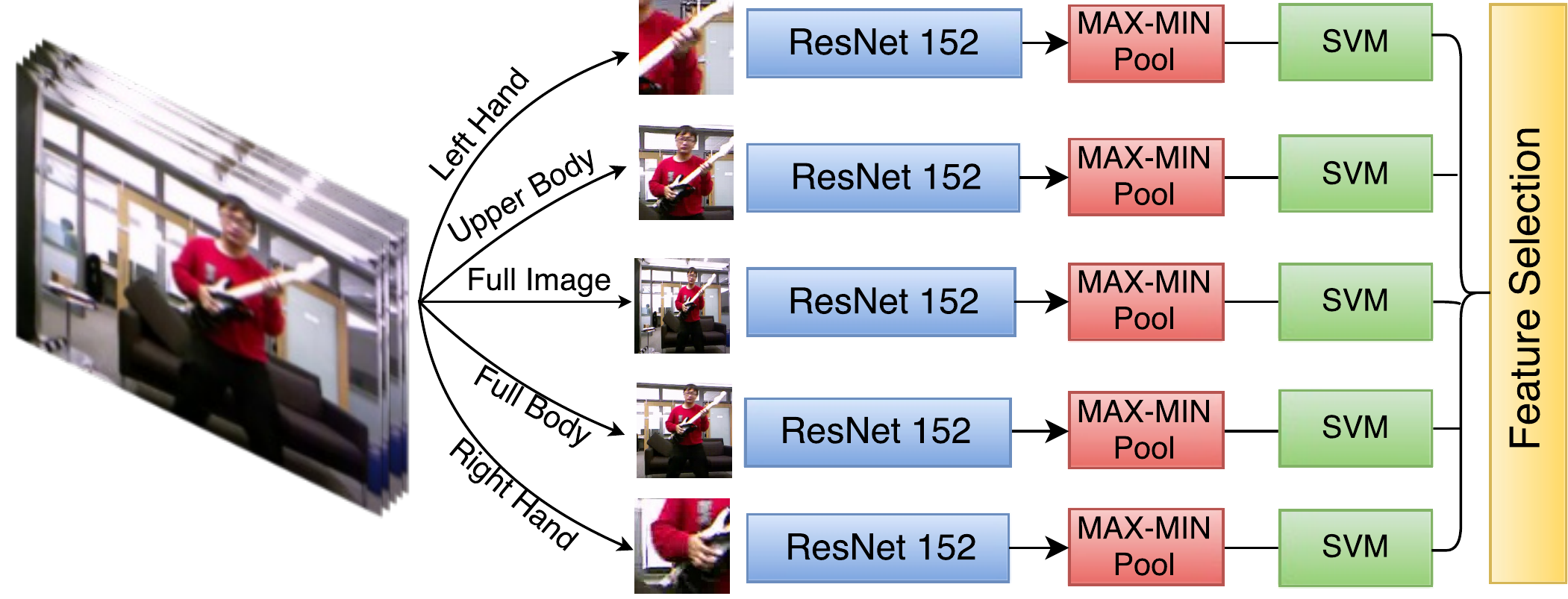} % include this on page two
\end{center}
   \caption{Each image frames are divided into five parts from their pose information which are input to ResNet-152 followed by max-min pooling. The classification from the SVM determines the part to be selected. The figure is only depicted for RGB stream and the Flow stream follows similar trend.}
\label{cnn_framework}
\end{figure}
In our pose based action network, CNN features from the left hand, right hand, upper body, full body and full images from each frame are extracted to represent each body region for the classification task as illustrated in fig.~\ref{cnn_framework}. Our experimental studies show that this body region representation leads to a lot of redundancy. Sometimes, wrong patches extracted due to side view actions which mislead the classifier to select a wrong action. Thus we propose a technique to select the best representation of the appearance feature by focusing on the body region with the most discriminative information. The patch representation for a given image-region $i$ is convolutional network $f_g$ with parameters $\theta_g$, taking as input a crop taken from image $I_t$ at the position of the part patch i:
\begin{equation}
z_{t,i} = f_g(crop(I_t,patch_i); \theta_g) \hspace{10mm} i=\{1,...,5\}
\end{equation}
We use pre-trained Resnet-152 for $f_g$ to extract the deep features from the last fully connected layer which yields 2048 values described as our frame descriptors $z_{t,i}$. These frame descriptors are aggregated over time using max and min pooling so as to focus the most salient values on the feature maps representing the video descriptors $z_{i}$. \\
The feature selection is done by feeding these CNN features $z_{i}$ to linear SVM for classification separately for each $i$. These SVM produce classification scores on cross validation set separately for each patch $i$. We select the patch $i$ of image-region with the best classification score on validation set. This allows us to select the best body region for characterizing the appearance feature. As per our observation, these selected appearance features not only represent the best static appearances but also have the best combinational power with the motion based information.\\

\vspace{-0.1in}
\subsection{Fusing Geometric and Appearance-Motion Features for Action Modeling}
 In this work, we propose to combine the discriminative power of 3D skeleton sequences with RGB based appearance and motion. Authors in~\cite{RNNandCNN,chainedcnn,twostream,twostreamfusion} attempts to fuse the appearance from RGB information, poses and motion. Here, the novelty lies in an attempt to fuse features from convolutional network and recurrent network in order to encode spatial and temporal features together.\\
The body based CNN features model the salient features in the global video. But these features are not discriminative enough to model the difference between actions with less intra-class variance. On the other hand, the features from LSTM models the temporal evolution of the salient features over the entire video. It captures the geometric evolution of the activity performed by a subject.   
So, the idea in this paper is to fuse the appearance based CNN features with the temporal evolution of the body translated skeleton joints. This is done by fusing the classification scores from SVM trained on deep appearance features and SVM trained on latent LSTM features. 
\vspace{-0.1in}
\section{Experiments}
\subsection{Dataset Description}
For evaluating our framework, we used \textbf{CAD-60}~\cite{40} containing 60 RGB-D videos with 14 actions performed by 4 subjects, \textbf{MSRDailyActivity3D}~\cite{8a} containing 320 RGB-D videos with 16 actions performed by 16 subjects and \textbf{NTU-RGB+D}~\cite{NTU_RGB+D} containing 56880 RGB-D videos with 60 actions performed by 40 subjects. \\
We evaluate CAD-60 and MSRDailyActivity3D by setting up a cross subject training and testing validation set up. For NTU-RGB+D, we follow the training/testing protocol mentioned in~\cite{NTU_RGB+D}. Our view transformation on 3D skeleton is performed to handle the side view actions performed by the subject on fixed camera, we are not focusing on Cross-View problem. Hence, we have not evaluated cross-view accuracy on NTURGB+D dataset.
\subsection{Implementation Details}
We build our LSTM framework on the platform of keras toolbox ~\cite{chollet2015keras} with TensorFlow~\cite{tensorflow2015}. The concept of Dropout~\cite{Dropout} is used with a probability of 0.5 to eliminate the problem of overfitting. The concept of Gradient clipping~\cite{gradientclipping} is used by restricting the norm of the gradient to not to exceed 1 in order to avoid the gradient explosion problem. Adam optimizer~\cite{adam_optimizer} initialized with learning rate 0.005 is used to train both networks. 
\vspace{-0.1in}
\subsection{Ablation Study}
In this section, we present the performance and analysis of the two cues used independently for action classification. Table~\ref{LSTM_comparison} shows the performance of different variants of LSTM with skeleton joints as input, used in the state-of-the-art. The performance of our proposed deep temporal LSTM feature extractor followed by linear SVM classifier outperforms the other LSTM variants. This is because of considering the predictions at each time step of the video sequence and classifying the latent representant of time for a video sample, which improves the temporal modeling of the classifier. This also opens a direction of using our approach with different input features and even with LSTM variants proposed in the state-of-the-art for LSTMs. \\
\begin{table}[!ht]
\centering
\begin{tabular}{|l|c|c|c|}
  \toprule
  Method & CAD60 & MSRDaily & NTU- \\
        &       & Activity3D & RGB+D  \\
  \midrule
  \small{Traditional LSTM} & 64.65 & 80.90 & 60.69 \\
  \small{Deep LSTM ~\cite{NTU_RGB+D}} & - & - & 60.7 \\
  \small{P-LSTM~\cite{NTU_RGB+D}} & - & - & 62.93\\
  \small{ST-LSTM (Joint-chain)~\cite{st-lstm}} & - & - & 61.7\\
  \textbf{Deep Temporal LSTM} & \textbf{67.64} & \textbf{91.56} & \textbf{64.49}  \\
  \bottomrule                             
\end{tabular}
\caption{Comparison of different approaches with body translated skeleton coordinates on CAD-60, MSRDailyActivity3D and NTU-RGB+D. The numbers here denote the accuracy [\%].}
\label{LSTM_comparison}
\end{table}

Table~\ref{Feature_selection} shows the effectiveness of our feature selection mechanism on different image regions for modeling the actions. It confirms the presence of irrelevant features (due to the combination of all body regions) which deviates the classifier decision boundary from the ideal one. This is also justified by the fact that some image regions may not be extracted correctly in some sequences and some may not have any significance in modeling the action. For instance, extracting the features from right hand for a person drinking water with left hand is of no significance. 

\begin{table}[!ht]
\centering
\begin{tabular}{|l|c|c|c|}
  \toprule
  Method & CAD60 & MSRDaily & NTU- \\
        &       & Activity3D & RGB+D  \\
  \midrule
  \small{P-CNN~\cite{22}} & 95.59  & 87.81 & 48.71 \\
  \hline
  \textbf{$FS$(P-CNN)} &  \textbf{97.06} & \textbf{89.06}  & \textbf{58.69}  \\
  \bottomrule                             
\end{tabular}
\caption{Effectiveness of pose based CNN features with feature selection mechanism on CAD60, MSRDailyActivity3D and NTU-RGB+D. The numbers here denote the accuracy [\%] and $FS$ corresponds to the feature selection mechanism.}
\label{Feature_selection}
\end{table}
\vspace{-0.1in}
\subsection{Comparison with the state-of-the-art}

Table~\ref{dataset_accuracy} presents the state-of-the-art comparison of our proposed fusion of depth based LSTM features and pose based CNN features to classify actions. The complementary nature of the LSTM and CNN based networks are evident from the boosted performance for MSRDailyActivity3D and NTU-RGB+D on fusion. The presence of static actions like \textit{cooking, talking on phone, relaxing on couch} and so on in CAD-60 do not enable the LSTM to recognize the dynamicity of the actions. This explains the gainless accuracy reported for CAD-60 on fusing the features. We outperform state-of-the-art results on CAD-60 and MSRDailyActivity3D.~\cite{chainedcnn} and~\cite{STA-hands} outperforms our proposed method using multi-stream 3D convolutions and attention mechanism respectively. Such mechanisms are hard to train and may not have consistent performance on smaller dataset. We observe that in NTU-RGB+D, short term motion is important which can be modeled using dense trajectory features~\cite{DT}(IDT-FV). Thus, we combine the IDT-FV features using a late fusion of individual classification score significantly boosting the performance over using individual features only and resulting in state-of-the-art performance on NTU-RGB+D. 

\begin{table}[!ht]
\begin{center}
\begin{tabular}{|l|c|}
\hline
Method & Accuracy [\%] \\
\hline\hline
Object Affordance ~\cite{35} & 71.40 \\
HON4D ~\cite{24} & 72.70 \\
Actionlet Ensemble ~\cite{8a} & 74.70 \\
MSLF ~\cite{33} & 80.36 \\
JOULE-SVM ~\cite{JOULE-SVM} & 84.10 \\
P-CNN + kinect + Pose machines ~\cite{Srijan} & 95.58 \\
\textbf{Proposed Method} & \textbf{97.06} \\
\hline
P-CNN + kinect + Pose machine ~\cite{Srijan} & 84.37 \\
Actionlet Ensemble ~\cite{8a} & 85.80 \\
RGGP + fusion ~\cite{29} & 85.60 \\
MSLF ~\cite{33} & 85.95 \\
DCSF + joint ~\cite{27} & 88.20 \\
JOULE-SVM ~\cite{JOULE-SVM} & 95.00 \\
Range Sample ~\cite{Range_sample} & 95.60 \\
DSSCA-SSLM ~\cite{MSRnew} & 97.50 \\
\textbf{Proposed Method} & \textbf{98.44} \\
\hline
$HOG^2$ ~\cite{hog2} & 32.2  \\
FTP DS ~\cite{JOULE-SVM} & 60.23 \\
Geometric features ~\cite{gemetricfeaturesWACV2017} & 70.26\\
Enhanced skeleton visualization ~\cite{liu2017enhanced}& 75.97\\
Ensemble TS-LSTM  ~\cite{lstm3d} & 74.60 \\
DSSCA-SSLM ~\cite{MSRnew} & 74.86\\
Chained Multistream Network ~\cite{chainedcnn} & 80.8 \\
STA-hands~\cite{STA-hands} & 82.5 \\
\textbf{Proposed Method} & \textbf{74.75} \\
\textbf{Proposed Method + IDT-FV} & \textbf{84.22} \\
\hline
\end{tabular}
\end{center}
\caption{Recognition Accuracy comparison for CAD-60 ($1^{st}$ section), MSRDailyActivity3D ($2^{nd}$ section) and NTU-RGB+D ($3^{rd}$ section) dataset. Proposed Method signifies Deep-Temporal LSTM + $
Feature Selection$ (P-CNN).}
\label{dataset_accuracy}
\end{table}
\vspace{-0.1in}
\section{Conclusion}
In this work, we propose a deep-temporal LSTM which models better temporal sequences as compared to the state-of-the-art architectures on the same input features and extended the pose based CNN action network by employing a feature selection mechanism. We also present the idea of fusing the pros of 3D skeleton based geometric features with appearance and motion based deep features to classify daily living activities. \\
A future direction lies in exploring different efficient features and variants of gating mechanism of LSTMs with our proposed approach. In the appearance-motion stream, the feature selection mechanism to select the appropriate image region is globally decided over the dataset. An attempt to select the appropriate feature for each sample and employing such a mechanism in the network itself is a direction to be explored.

{\small
\bibliographystyle{ieee}
\bibliography{egbib}
}

\rule{0pt}{1pt}\newpage
\rule{0pt}{1pt}\newpage
\rule{0pt}{1pt}\newpage
\rule{0pt}{1pt}\newpage
\rule{0pt}{1pt}\newpage

\end{document}